\documentclass{article}

\PassOptionsToPackage{numbers,square}{natbib}
\usepackage[final]{nips_2016}

\usepackage[utf8]{inputenc} 
\usepackage[T1]{fontenc}    
\usepackage{hyperref}       
\usepackage{url}            
\usepackage{booktabs}       
\usepackage{amsfonts}       
\usepackage{nicefrac}       
\usepackage{microtype}      
\usepackage{color}
\usepackage{courier}

\usepackage[pdftex]{graphicx}

\usepackage{listings}

\definecolor{codegreen}{rgb}{0,0.3,0}
\definecolor{codeblue}{rgb}{0,0.0,0.5}
\definecolor{codegray}{rgb}{0.5,0.5,0.5}
\definecolor{codebackground}{rgb}{0.98,0.98,0.98}

\lstdefinestyle{codestyle}{
	language=Python, 
	backgroundcolor=\color{codebackground},   
	commentstyle=\color{codegreen},
	keywordstyle=\color{codeblue},
	numberstyle=\tiny\color{codegray},
	stringstyle=\color{codegreen},
	basicstyle=\footnotesize\ttfamily,
	frame=none, 	
	breakatwhitespace=false,         
	breaklines=true,                 
	captionpos=b,                    
	keepspaces=true,                 
	numbers=left,                    
	numbersep=5pt,                  
	showspaces=false,                
	showstringspaces=false,
	showtabs=false,                  
	tabsize=2,
	morekeywords={BuildBatch, Map, Collect, MapCol, Zip, ReadPandas, SplitRandom, Stratify, ReadImage,
	              KerasNetwork, LogToFile, AugmentImage, TransformImage, Consume, Range, Filter, Take},  
}

\title{nuts-flow/ml : data pre-processing for deep learning}

\newcommand*{\affaddr}[1]{#1} 
\newcommand*{\affmark}[1][*]{\textsuperscript{#1}}
\newcommand*{\email}[1]{\texttt{#1}}

\author{%
	S. Maetschke\affmark[1], R. Tennakoon\affmark[2], C. Vecchiola\affmark[1] and R. Garnavi\affmark[1]\\
	\\
	\affaddr{\affmark[1]IBM Research Australia, 5/204 Lygon Street, Melbourne VIC 3053}\\
	\affaddr{\affmark[2]School of Engineering, RMIT University, 376 Swanston Street, Melbourne VIC 3000}\\
	\email{\{stefanrm,christian.vecchiola,rahilgar\}@au1.ibm.com}\\
	\email{ruwan.tennakoon@rmit.edu.au}\\%
}

\begin{document}

\maketitle

\begin{abstract}
Data preprocessing is a fundamental part of any machine learning application and frequently the most time-consuming aspect when developing a machine learning solution. Preprocessing for deep learning is characterized by pipelines that lazily load data and perform data transformation, augmentation, batching and logging. Many of these functions are common across applications but require different arrangements for training, testing or inference. 
Here we introduce a novel software framework named \emph{nuts-flow/ml} that encapsulates common preprocessing operations as components, which can be flexibly arranged to rapidly construct efficient preprocessing pipelines for deep learning.
\end{abstract}

\section{Introduction}

Human level performance and beyond have been achieved in many vision and other machine learning applications due to deep learning \cite{He2015}. Deep learning describes a class of artificial neural networks characterized by many layers and weights \cite{Krizhevsky2012}. The large number of parameters to be optimized during training requires large amounts of data and hardware accelerators such as  Graphical Processing Units (GPUs) or dedicated hardware such as the Tensor Processing Unit (TPU) \cite{Jouppi2017} for tensor computations. Training is typically performed via  Stochastic Gradient Descent (SGD) or variants thereof \cite{Ruder2016} which adjust network weights based on small batches (mini-batches) of data.

\begin{figure}[h]
  \centering
  \includegraphics[scale=0.6]{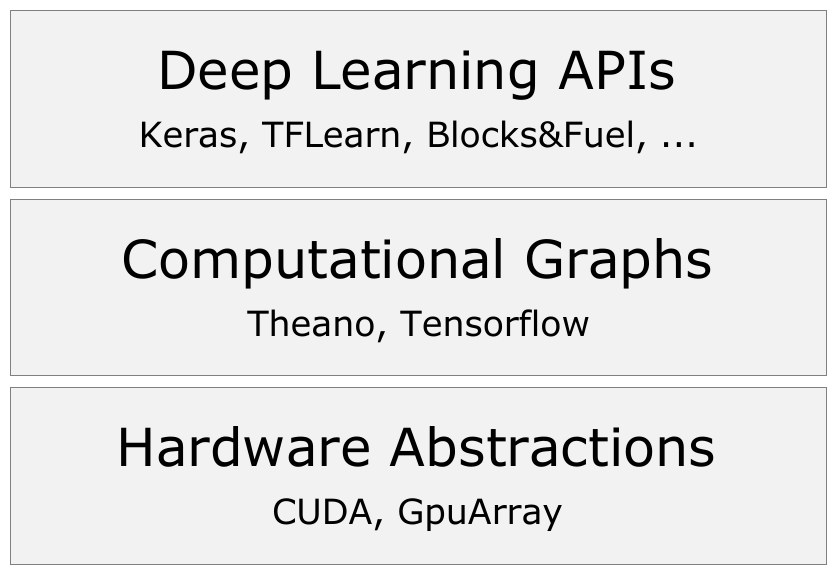}
  \caption{Deep Learning stack.}
  \label{fig:stack}
\end{figure}

Deep learning frameworks \cite{Bahrampour2016} are composed of up to three software layers (see Figure \ref{fig:stack}). On the lowest level, are hardware abstractions such as CUDA \cite{CUDA} or GpuArray \cite{GpuArray}. The middle layer is focused on the manipulation of tensors and the construction of computational graphs that are employed to create and train networks. Theano \cite{Theano2016} and Tensorflow \cite{Abadi2016} are examples of such libraries or backends. The top layer consists of APIs to simplify the definition of architecture, weight initializations, loss functions and learning algorithms for deep learning networks. For instance, Keras \cite{Chollet2015}, TFLearn \cite{Tang2016}, Blocks\&Fuel \cite{Merrienboer2015} and Caffe \cite{Jia2014} are libraries with Python APIs used for this purpose. Torch \cite{Collobert2011} based on Lua, Mocha \cite{Mocha} based on Julia, and Deeplearing4J \cite{DL4J} based on Java are common non-Python alternatives. We will focus our discussion on the former, since the framework presented here is based on Python.

Data preprocessing is usually performed within the top or middle layer. But while the above-mentioned libraries greatly reduce the effort to construct deep learning networks, they lack similar support for data-preprocessing such as lazy loading, filtering and transforming of input data.
However, data-preprocessing is a fundamental part of any machine learning task and often the most time-consuming during development and performance tuning. Similar to the iterative refinement of network hyper-parameters such as architecture, learning rate, loss function and others, the implementation of the data-preprocessing step is also an iterative process which involves, for instance, the exploration of different data normalizations, augmentations or filtering procedures.

Data preprocessing for deep learning is especially challenging due the following characteristics, 1) training data cannot be loaded in memory entirely and must be processed lazily, 2) the training set is often enriched by random augmentation, 3) training is based on batches and 4) some data preprocessing occurs on the CPU but training and inference are performed by the GPU.

In the following sections, we first discuss the preprocessing capabilities of existing deep learning frameworks, then describe the foundations of data preprocessing pipelines, and finally introduce the novel data preprocessing framework \emph{nuts-flow/ml}, before closing with conclusions.

\section{Background}

Existing deep learning frameworks are focused on the definition and training of artificial neural networks but provide no or very limited support for data preprocessing. They largely rely on NumPy~\cite{NumPy}, Scikit~\cite{Pedregosa2011}, pandas~\cite{McKinney2011} and similar deep-learning-agnostic libraries, which are designed for in-memory processing and have limited support for lazy data flows as required for deep learning. Some frameworks, however, have built-in preprocessing functionality; in the following we discuss Keras, TFLearn, Tensorflow, Caffe and Fuel in more detail.

Keras provides an {\tt ImageDataGenerator} for standard augmentations such as rotation or flipping of images, but is not easily extensible to other augmentations, does neither support patch generation nor allows applying the same random augmentation to two images at the same time, e.g. image and mask, which is a common requirement for segmentation tasks. An important feature of Keras, is that it performs pre-fetching of data; allowing data to be preprocessed on the CPU, while another data chunk is evaluated for training or inference on the GPU.

TFLearn has {\tt DataPreprocessing} and {\tt DataAugmentation} classes that can be extended by custom preprocessing functions. They operate on batches, which results in more efficient but also more difficult to write code, e.g. applying random augmentations to multiple images synchronously or patch generation are challenging to implement. However, TFLearn provides a {\tt DataFlow} class to construct concurrent data flows that goes beyond what nuts-flow/ml offers with respect to concurrency.

Tensorflow comes with an {\tt image} module that implements a rich set of image transformations and random augmentations. Adding custom transformations is straight forward, though the actual coding is not trivial since it requires manipulation of batched data. There is no built-in support for random patch extraction or patch extraction in regions of interest.

Caffe offers a {\tt Transformer} class with a small set of image transformations and a {\tt Data/Python Layer} to implement custom transformation and augmentation. However, implementing custom data layers is not an easy task. There are add-on libraries such as caffe-augmentation \cite{CaffeAugmentation} that support a wider range of augmentations but the functionality is still limited and patching is not included.

The Fuel library is closest in spirit and functionality to the nuts-ml framework presented here. It allows the construction of data flows with a focus on image/text preprocessing and deep-learning. Custom transformations and augmentations can be added but their implementation requires considerably more code than in nuts-flow/ml (see example below). Patching or synchronized augmentations are not directly supported. Syntactically, flows are defined as nested function compositions, resulting in comparatively verbose and difficult-to-read pipelines.

The following excerpt from the Fuel tutorial \cite{FuelExample} shows a (toy) data flow for batch generation. For brevity import statements are omitted. The flow takes a list of feature values {\tt [1, 2, 3, 4]} and a list of target values {\tt [-1, 1, -1, 1]}, constructs a dataset ({\tt IndexableDataset}), multiplies all target values by 2 ({\tt Doubler}) and returns an iterator ({\tt DataStream}) over batches of size 2 ({\tt SequentialScheme}):

\begin{lstlisting}
>>> class Doubler(AgnosticSourcewiseTransformer):
...     def __init__(self, data_stream, **kw):
...         super(Doubler, self).__init__(
...             data_stream=data_stream,
...             produces_examples=data_stream.produces_examples, **kw)
...
...     def transform_any_source(self, source, _):
...         return 2 * source

>>> dataset = IndexableDataset(
...     indexables=OrderedDict([
...         ('features', np.array([1, 2, 3, 4])),
...         ('targets', np.array([-1, 1, -1, 1]))]))

>>> batch_scheme = SequentialScheme(
...     examples=dataset.num_examples, batch_size=2)

>>> target_stream = Doubler(
...     data_stream=DataStream(
...         dataset=dataset,
...         iteration_scheme=batch_scheme),
...     which_sources=('targets',))

>>> [batch for batch in target_stream.get_epoch_iterator()]
[(array([1, 2]), array([-2,  2])), (array([3, 4]), array([-2,  2]))]
\end{lstlisting}

In short, feature and target values are zipped, target values are doubled and mini-batches are generated. The same pipeline can be realized in three lines (again omitting imports) using nuts-flow/ml:

\begin{lstlisting}
>>> dataset = [1, 2, 3, 4] >> Zip([-1, 1, -1, 1])
>>> build_batch = BuildBatch(2).by(0, 'number', int).by(1, 'number', int)
>>> dataset >> MapCol(1, _ * 2) >> build_batch >> Collect()
[[array([1, 2]), array([-2,  2])], [array([3, 4]), array([-2,  2])]]
\end{lstlisting}

Not only is the resulting code shorter and more readable but also the direction of the data flow is clearly indicated by the '{\verb >> }' operator. We provide a detailed explanation of the above code elements in a later section.

\emph{NumPy} is a library that is frequently employed for data pre-processing and nuts-flow/ml is using NumPy internally. The approximately equivalent code of the example above using plain \emph{NumPy} is only marginally shorter but not lazy and does not perform pre-fetching. The {\tt split()} function constructs all mini-batches at once in memory, which is infeasible for large datasets.

\begin{lstlisting}
>>> dataset = np.array([[1, 2, 3, 4], [-1, 1, -1, 1]])
>>> dataset[1,:] *= 2
>>> [list(b) for b in np.split(dataset, 2, 1)]
[[array([1, 2]), array([-2,  2])], [array([3, 4]), array([-2,  2])]]
\end{lstlisting}

Fuel, nuts-ml and other preprocessing front-ends discussed above are designed for rapid prototyping of small to medium sized data sets. Large scale, cluster computing frameworks such as \emph{Apache Spark} \cite{Zaharia2016} or \emph{Das}k \cite{Dask2016} also support lazy evaluation and have a concept of data flows. However, these frameworks are heavy-weight, have high latencies, do not directly provide preprocessing functions for images or deep learning and generally are not suitable for rapid prototyping of deep learning systems. It is worth mentioning that flows implemented with nuts-ml can be broken into their components or sub-flows and integrated within Spark or Dask workflows.

To summarize: current data preprocessing implementations for deep learning generally support only basic transformations and augmentations but common, more complex use cases, such as patching, or simultaneous random augmentation and transformation of multiple images, e.g. for segmentation are not directly available. Furthermore, extending existing frameworks to implement missing functionality is often challenging and the resulting pipelines tend to lack readability. For instance, the example pipeline discussed above, requires more than 10 lines of heavily nested code using Fuel, while the same functionality can be implemented in three, easily understandable lines using nuts-flow/ml. Also support for image processing is very limited or non-existent in most pipelines, while nuts-flow/ml offers a rich set functions specifically for this purpose.

In the following section, we discuss data preprocessing pipelines and common design pattern for their implementation within a Python environment in general, before describing the specifics of the nuts-flow/ml implementation.

\section{Data processing pipelines}

\begin{figure}[h]
	\centering
	\includegraphics[scale=0.55]{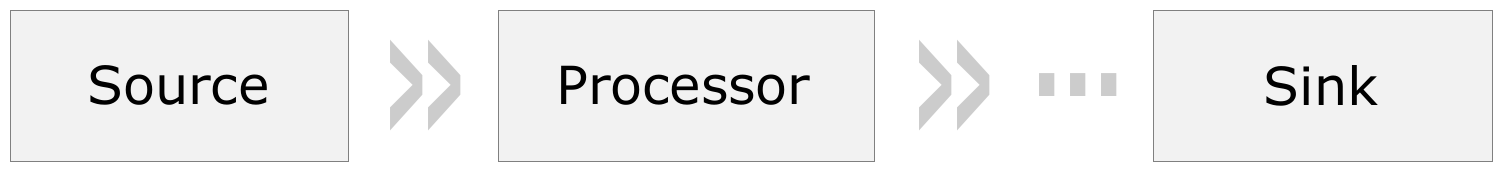}
	\caption{Data processing pipeline.}
	\label{fig:procpipe}
\end{figure}

Data processing pipelines read data from a \emph{Source}, transform data using \emph{Processors}, and finally write data to a \emph{Sink} (see Figure \ref{fig:procpipe}). The main advantage and purpose of a processing pipeline compared to other programmatic constructs is the  \emph{lazy} evaluation of data. Only a subset of the entire data volume occupies memory at a time and data is processed on demand. This enables the efficient processing of data too large for a computer’s main memory including infinitely large data volumes.

Data processing pipelines, specifically for deep learning, are characterized by a common sequence of steps that can be summarized in a \emph{Canonical Pipeline}, described in the next section.

\subsection{Canonical Pipeline}

\begin{figure}[h]
	\centering
	\includegraphics[scale=0.6]{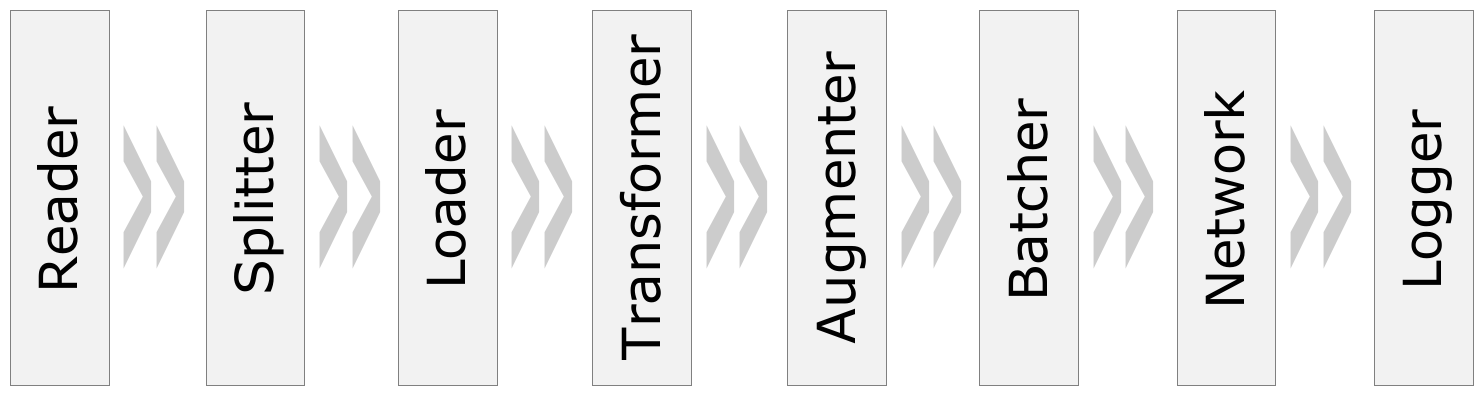}
	\caption{Canonical data preprocessing pipeline.}
	\label{fig:canpipe}
\end{figure}

Deep learning pipelines read samples, split data sets into fold, load images or other large data, apply transformation and augmentations, build mini-batches, before feeding a network with data and logging training results. These common processing steps can be represented in a \emph{Canonical Pipeline} (see Figure \ref{fig:canpipe}). For simplicity, we focus the following more detailed description on vision tasks and image data but any deep learning application with data too large to fit into memory such as video, audio or large text documents, will benefit from similar functionality.

The \emph{Reader} component of the pipeline reads sample data, for instance, paths to image files and class labels, stored in text files, Pandas tables or databases. The sample set is then partitioned into training, validation and test folds by a \emph{Splitter}. Since the training data cannot be loaded into memory completely, images or other data blobs are loaded lazily by a \emph{Loader}. Typically, transformations of image data such as resizing, cropping, contrast normalization and other adjustments are needed and performed on-the-fly by a \emph{Transformer}. Furthermore, to increase the training set, additional images are synthesized by randomly augmenting (rotating, flipping, …) training images employing an \emph{Augmenter}. Efficient, GPU-based training demands that data is organized in small batches by a \emph{Batcher} before passed on to the Network for training or inference. Finally, the training progress is often monitored using a \emph{Logger} that writes losses or accuracies to a log file.

It is worth reiterating that all preprocessing steps that operate on large data such as loading, transformation, augmentation and batching need to be performed lazily, avoiding the allocation of main memory for data that is not actively processed. 

\subsection{Iterators and itertools}

A common design pattern to implement lazy evaluation is the \emph{Iterator} \cite{Kasampalis2015}, which ensures that data is processed on demand only. Pipelines in nuts-flow/ml are implemented as chains of iterators. Here we, firstly, discuss Python’s iterator construct and its iterator library, before describing their specific usage in the architecture of nuts-flow/ml.

Iterators in the Python programming language (Version 2.7) are classes that expose a {\tt next()} 
\footnote{ Python 3.x replaces {\tt next()} with {\tt \_\_next\_\_()}. }
method. Each invocation of {\tt next()} returns the next element of the iterator. For instance, the following class implements an iterator that generates even numbers:

\begin{lstlisting}
class EvenNumbers():
    def __init__(self):
        self.number = 0
        
    def next(self):
        self.number += 2
        return self.number
        
    def __iter__(self):
        return self
\end{lstlisting}

The {\verb __iter__() } method enables the iterator to be used in for-loops. The following code example would print even numbers infinitely:

\begin{lstlisting}
for e in EvenNumbers():
    print e  # 2, 4, 6, ...
\end{lstlisting}

Iterators can be chained and transformed to construct lazy data processing pipelines. Python’s {\tt itertools} library provides many common iterators and functions for this purpose. The following example demonstrates a simple pipeline that generates number from 0 to 9, filters all numbers greater than 5, takes the first 3 numbers, and collects the resulting numbers in a list:

\begin{lstlisting}
list(islice(ifilter(lambda x: x > 5, xrange(10)), 3))
\end{lstlisting}

Every iterator pipeline requires a sink, here {\tt list()}, that drives the data flow by repeatedly calling {\tt next()} until the data source, here {\tt xrange()}, is depleted, or until an intermediary iterator terminates. In this case {\tt islice()} terminates after 3 elements are taken from the data flow.

Python’s iterators and the {\tt itertools} library allow constructing efficient and lazy data flows. However, the nested syntax of calls results in pipelines that are difficult to understand and the library has no functionality specific to image processing and machine learning tasks. In the following sections, we introduce two novel libraries, \emph{nuts-flow} and \emph{nuts-ml}, that are based on {\tt itertools} but describe data flows as a linear sequence of operations and provide functions that greatly simplify data preprocessing for deep learning.

\section{nuts-flow/ml}

\emph{nuts-flow/ml} is composed of \emph{nuts-flow} and \emph{nuts-ml}. The former is a general-purpose data flow library in spirit of the functional programming paradigm and not specific to any application. It is lightweight and does not have any dependencies beyond Python and its standard libraries. The latter is an extension of nuts-flow, which adds application-specific functions for image processing and deep learning.

\subsection{nuts-flow}

Before discussing the details of nuts-flow we want to motivate its design and choice of syntax by re-implementing the simple {\tt itertools} pipeline shown above and comparing it with the corresponding nuts-flow implementation (imports are omitted):

\begin{lstlisting}
# itertools
list(islice(ifilter(lambda x: x > 5, xrange(10)), 3))
\end{lstlisting}

\begin{lstlisting}
# nuts-flow
Range(10) >> Filter(_ > 5) >> Take(3) >> Collect()
\end{lstlisting}

Both implementations return the exact same result: a list with the numbers {\tt [6, 7, 8]}. However, the nuts-flow code is easier to understand and the flow of data from left to right is clearly visible. The syntax is similar to many other language operators for function composition or data flows such as Unix pipes '{\verb | }', C++ streams '{\verb >> }', Haskell’s function composition '{\tt \$}' or dplyr’s chaining of operators '{\tt \%>\%}', For nuts-flow we overload Python’s '{\verb >> }' operator, which usually performs a right bit shift, since it is rarely used in the applications we are interested in and visually indicates the direction of the data flow.

\subsubsection{Architecture}

Data flows in nuts-flow are implemented as chains of processing components called \emph{nuts}. With some exceptions discussed later, nuts take an iterable as input and return a Python iterator or generator. In addition, nuts can have parameters that control their function. All nuts are class objects derived from the following abstract base class:

\begin{lstlisting}
class Nut():

    def __rshift__(self, iterable):
        pass
\end{lstlisting}

Custom nuts need to override the special {\verb __rshift__() } function \cite{PythonSpecialFunctions}, which represents the '{\verb >> }' operator. An example of a custom nut that multiplies its input values by a given factor is shown below:

\begin{lstlisting}
class MultiplyBy(Nut):

    def __init__(self, factor):
        self.factor = factor

    def __rshift__(self, iterable):
        return (x * self.factor for x in iterable)
\end{lstlisting}

This custom nut could then be called in a data flow as follows, where {\tt Collect()} collects the outputs of the nut in a list:

\begin{lstlisting}
>>> [1, 2, 3] >> MultiplyBy(2) >> Collect()
[2, 4, 6]
\end{lstlisting}

Note that this is equivalent to the following code that is considerably less readable due to the nesting of function invocations. Using the '{\verb >> }' operator syntax instead, untangles the nested function composition to a linear chain of processing steps.

\begin{lstlisting}
Collect().__rshift__(MultiplyBy(2).__rshift__([1, 2, 3]))
\end{lstlisting}

The nuts-flow library provides wrapper classes and decorators to simplify the definition of custom nuts and so facilitates the extension of the framework. For example, the {\tt MultiplyBy()} nut could more succinctly be defined as follows,

\begin{lstlisting}
@nut_function
def MultiplyBy(x, factor):
    return x * factor
\end{lstlisting}

where the decorator {\verb @nut_function } transforms the decorated function to a class implementation similar to the one shown in the example above.

All data flows require a source that emits data and a sink that consumes data and drives the flow. Between source and sink there is typically a sequence of processors such as functions or filters that transform the data flow (see Figure \ref{fig:procpipe}).

Sinks call {\tt next()} on their inputs that are generated by processors, which themselves invoke {\tt next()} on their inputs and so create a chain of iterator calls that lazily read data from the source. Without a sink, the flow is inactive and no data is processed.

Since nuts flows are chained iterators without an underlying workflow engine, they do not automatically support concurrency. They are also not well suited to describe complex data flows that take the shape of directed acyclic graphs, where processing nodes can have multiple inputs and outputs that must be synchronized. However, data preprocessing pipelines for deep learning tend to be largely linear and individual processing steps can be parallelized within nuts-flow. 

The restriction to iterator chains has the advantage of a simple implementation and enables the easy integration with plain Python or cluster frameworks. For instance, the following three examples are alternative implementations of the flow described above and demonstrate seamless integration with Python:

\begin{lstlisting}
>>> xrange(1, 4) >> Map(_ * 2) >> Collect()
[2, 4, 6]
\end{lstlisting}

\begin{lstlisting}
>>> list(xrange(1, 4) >> MultiplyBy(2))
[2, 4, 6]
\end{lstlisting}

\begin{lstlisting}
>>> [MultiplyBy(2)(x) for x in [1, 2, 3]]
[2, 4, 6]
\end{lstlisting}

In addition to functional programming operations such as {\tt map}, {\tt filter} and {\tt reduce} that are native to Python, and the iterator functions in {\tt itertools}, nuts-flow provides many more nuts to construct flows. A description of the complete set is beyond the scope of this paper but noteworthy are nuts for reading and writing of CSV files, for printing and progress monitoring, for pre-fetching, caching and parallel processing, for exception handling, for chunking and grouping of data and for column-wise mapping of functions \cite{NutsflowOverview}. 

\subsection{nuts-ml}

\begin{figure}[h]
\centering
\includegraphics[scale=0.6]{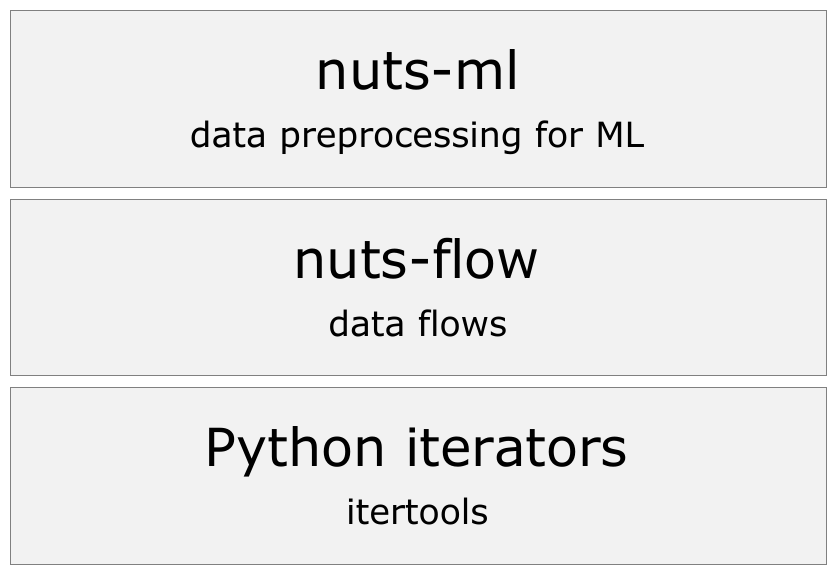}
\caption{Architecture.}
\label{fig:architecture}
\end{figure}

\emph{nuts-ml} is based on \emph{nuts-flow} but adds nuts for image preprocessing and machine learning, while \emph{nuts-flow} is an extension of Python’s {\tt itertools} library to implement generic data flows. Figure \ref{fig:architecture} depicts the overall architecture of the library.

In the following, we will first describe individually the nuts available in nuts-ml that serve as components of the Canonical Pipeline, and then demonstrate a complete pipeline. Assuming a data set {\tt data.csv} is given as Pandas table (or CSV file) and contains data samples of the form {\tt (filepath, label)}, where {\tt filepath} points to an image file and {\tt label} is a class label:

\begin{lstlisting}
data.csv
'image1.png', 'plane'
'image2.png', 'car'
...
\end{lstlisting}

Such a dataset can be read with the Pandas reader nut that returns an iterator over the samples. The following example shows a very short pipeline where the samples are collected in a list using the {\tt  Collect} nut:

\begin{lstlisting}
>>> dataset = ReadPandas('data.csv') >> Collect()
>>> print dataset
[('image1.png', 'plane'), ('image2.png', 'car'), ...]
\end{lstlisting}

After reading sample data the training, hyper-parameter optimization and performance evaluation of deep learning networks typically requires different data folds for training, validation and testing. The {\tt SplitRandom} nut provides this functionality and divides a data set with given ratios. For instance, a 60\% training, 20\% validation and \%20 test data split is common:

\begin{lstlisting}
trainset, valset, testset = dataset >> SplitRandom(ratio=(60,20,20))
\end{lstlisting}

In most practical applications, the class distribution within the data set is skewed. Data with balanced class frequencies, however, tend to result in better training and generalization and typically the class distribution of the training set is therefore stratified, ensuring equal numbers of samples for all classes. Stratification is based on the sample column that contains the class label and up-sampling or random down-sampling are supported:

\begin{lstlisting}
trainset = trainset >> Stratify(labelcol=1, mode='up')
\end{lstlisting}

After splitting and stratification, the data set still contains filenames only but no actual images. The {\tt ReadImage} nut reads image data from the given {\tt imagepath} and replaces the filename in the given sample {\tt columns} 
\footnote{ {\tt ReadImage} can read multiple images and replace filepaths in multiple columns at once. }
by the image:

\begin{lstlisting}
imgsamples = trainset >> ReadImage(columns=0, imagepath='images/*')
\end{lstlisting}

As shown in Figure \ref{fig:canpipe}, part of the Canonical Pipeline is a \emph{Transformer} that converts images to a format suitable for training or inference. Typical transformations include resizing images to a specific size, image cropping, or converting RGB images to gray-scale. The {\tt TransformImage} nut transforms images in the given sample columns ({\tt imagecols}) by applying the specified transformations in the defined order:

\begin{lstlisting}
imgtrans = (imgsamples >> TransformImage(imagecols=0)
                          .by('resize', 128, 128)
                          .by('rgb2gray'))
\end{lstlisting}

Note that transformations can be chained (here resizing and gray-scale conversion) and can be applied to multiple images of a sample simultaneously, e.g. image and mask or stereo images to name some common use cases. Also, user-defined transformation can be added easily by registering transformation functions.

The training of deep learning models usually requires large amounts of training data. A common strategy to enlarge the existing training set is through augmentation, for instance, by randomly flipping or rotating images. The {\tt AugmentImage} nut takes images from the given sample columns and with given probability (here {\tt 0.5}) augments images:

\begin{lstlisting}
imgaug = (imgtrans >> AugmentImage(imagecols=0)
                      .by('fliplr', 0.5)
                      .by('rotate', 0.5, [0, 360]))
\end{lstlisting}

Similar to transformations, augmentations can be applied synchronously to multiple images within a sample and user defined augmentations can be added.

Efficient network training on GPUs demands the batching of training data. {\tt BuildBatch} stacks multiple images to build tensors and can also perform one-hot encoding of class labels to construct training batches. Analogous to transformations and augmentations, data is extracted from given sample columns and converted into required data types. In the following example, batches of size {\tt 16} are created, where images in sample column {\tt 0} are converted to unsigned integer format, and class label in column {\tt 1} are encoded as one-hot vectors of length {\tt 10}:

\begin{lstlisting}
batches = (imgaug >> BuildBatch(batchsize=16)
                     .by(0, 'image', np.uint8, True)
                     .by(1, 'one_hot', np.uint8, 10)))
\end{lstlisting}

Most APIs for deep learning frameworks provide methods to train on batches. For instance, the batches generated by {\tt BuildBatch} could directly be fed into a Keras model via {\verb model.fit_generator() }. nuts-ml provides wrappers for Keras and Lasagne models that simplify network training, inference and evaluation further. For example, a given Keras model can be wrapped into a {\tt KerasNetwork} nut and then trained on a flow of batches as follows:

\begin{lstlisting}
losses = batches >> KerasNetwork(model).train()
\end{lstlisting}

The {\tt train()} method returns an iterator over batch losses, which can continuously be written to a log file using a {\tt LogToFile}:

\begin{lstlisting}
losses >> LogToFile('losses.log') >> Consume()
\end{lstlisting}

or accumulated to the mean value and standard deviation via:

\begin{lstlisting}
loss, std = losses >> MeanStd()
\end{lstlisting}

Above we have introduced the individual nuts that are part of a typical deep learning pipeline. In the following final example, we demonstrate how to connect these nuts to form a fully functional pipeline. First, we re-define the nuts more succinctly and assign them to variables:

\begin{lstlisting}
read_data = ReadPandas('data.csv')
split_data = SplitRandom(ratio=(60,20,20))
stratify = Stratify(labelcol=1, mode='up')
read_images = ReadImage(0, imagepath='images/*')
transform = TransformImage(0).by('resize', 128, 128).by('rgb2gray')
augment = AugmentImage(0).by('fliplr', 0.5).by('rotate', 0.5, [0, 360])
network = KerasNetwork(model)
log = LogToFile('losses.log')
\end{lstlisting}

In the next step, we split the dataset into folds
\footnote{ Note that this is a non-lazy operation but computationally inexpensive since samples at this stage contain file paths only but no actual image data. }:

\begin{lstlisting}
trainset, valset, testset = read_data >> split_data
\end{lstlisting}

Training on the stratified samples can now be performed in a loop over epochs using the following pipeline, where {\tt Consume} is a nut that consumes all data and drives the data flow:

\begin{lstlisting}
for epoch in EPOCHS:
    (trainset >> stratify >> read_images >> transform >> augment >> 
     network.train() >> log >> Consume())
\end{lstlisting}

Note that all steps in this pipeline are performed lazily. Only images required to build one batch at a time are read, transformed and augmented. Once the nuts are defined it is trivial to construct similar pipelines for validation or network evaluation \cite{NutsmlCIFAR10}. In addition to the nuts shown here, nuts-ml provides nuts for image patching, generation of masks, display of images, boosting of samples, reading from label-directories and more \cite{NutsmlOverview}.

\section{Conclusion}

Data preprocessing is an essential part of any deep learning application. Existing APIs or libraries for preprocessing, however, are limited in functionality and often difficult to extend. Here, we introduced a novel software framework \emph{nuts-flow/ml} that encapsulates common preprocessing functions as components that can flexibly be arranged to rapidly construct efficient data preprocessing pipelines. 

nuts-flow/ml are well tested and documented libraries for Python 2.7 and 3k that considerably simplify the construction and rapid modification of data preprocessing pipelines for deep learning. Both libraries are freely available on github  \cite{NutsflowCode,NutsmlCode} under the Apache~2 license.

Future work will be focused on extending the library with preprocessing functionality for other data types than images such as audio, video and text. We also intend to provide additional network wrappers for other deep learning frameworks beyond Keras, with Theano or Tensorflow backend, and Lasagne, which are already supported.


\medskip
\small

\bibliographystyle{plainnat}
\bibliography{nutsml}

\end{document}